\documentclass{article}

\PassOptionsToPackage{numbers, compress}{natbib}
\usepackage[final]{neurips_2024}
\usepackage{color}
\usepackage[dvipsnames]{xcolor}
\usepackage{colortbl}
\usepackage{graphicx}
\usepackage{booktabs}
\usepackage{xspace}
\usepackage[utf8]{inputenc} 
\usepackage[T1]{fontenc}    
\usepackage[colorlinks,
            linkcolor=VioletRed,      
            anchorcolor=VioletRed, 
            citecolor=VioletRed,       
            ]{hyperref}
\usepackage{subfigure}
\usepackage{tcolorbox}
\usepackage{wrapfig}
\usepackage{url}            
\usepackage{booktabs}       
\usepackage{amsfonts}       
\usepackage{nicefrac}       
\usepackage{microtype}      
\usepackage{xcolor}         
\usepackage{bm}
\usepackage{graphicx}
\usepackage{amsmath}
\usepackage{cleveref}
\usepackage{amsthm}
\usepackage{tikz}
\usepackage{multirow}
\usepackage{graphicx}
\definecolor{Gray}{gray}{0.85}
\newcommand{\Gray}[0]{\rowcolor{gray!20}}

\definecolor{sclgreyblue}{rgb}{0.2,0.3,0.5}%

\usepackage{soul}

\usepackage{xcolor}
\definecolor{CFE2F3}{HTML}{CFE2F3}

\title{VITA: Towards Open-Source Interactive \\ Omni Multimodal LLM}
\vspace{-0.6cm}
\author{
\vspace{-0.4cm}
\\ 
    Chaoyou Fu$^{1,\spadesuit}$, Haojia Lin$^{3}$, Zuwei Long$^{2}$, Yunhang Shen$^{2}$, Yuhang Dai$^{2}$, Meng Zhao$^{2}$ 
    \\
    Yi-Fan Zhang$^{4}$, Shaoqi Dong$^{1}$, Yangze Li$^{2}$, Xiong Wang$^{2}$, Haoyu Cao$^{2}$, Di Yin$^{2}$, Long Ma$^{2}$
    \\
    Xiawu Zheng$^{3}$, Rongrong Ji$^{3}$, Yunsheng Wu$^{2}$, Ran He$^{4,\dagger}$, Caifeng Shan$^{1,\dagger}$, Xing Sun$^{2,\dagger}$
    \\
    $^{1}$NJU, $^{2}$Tencent Youtu Lab, $^{3}$XMU, $^{4}$CASIA
    \and
    \footnotesize{
    $^{\spadesuit}$~Project Leader \;
    $^{\dagger}$~Corresponding Author \;}
    \\ \\
    \url{https://vita-home.github.io}
}

\begin{document}

\maketitle

\vspace{-0.6cm}
\begin{abstract}
The remarkable multimodal capabilities and interactive experience of GPT-4o underscore their necessity in practical applications, yet open-source models rarely excel in both areas.
In this paper, we introduce \textbf{VITA}, the first-ever open-source Multimodal Large Language Model (MLLM) adept at simultaneous processing and analysis of \textbf{V}ideo, \textbf{I}mage, \textbf{T}ext, and \textbf{A}udio modalities, and meanwhile has an advanced multimodal interactive experience.
Starting from Mixtral 8$\times$7B as a language foundation, we expand its Chinese vocabulary followed by bilingual instruction tuning.
We further endow the language model with visual and audio capabilities through two-stage multi-task learning of multimodal alignment and instruction tuning.
VITA demonstrates robust foundational capabilities of multilingual, vision, and audio understanding, as evidenced by its strong performance across a range of both unimodal and multimodal benchmarks.
Beyond foundational capabilities, we have made considerable progress in enhancing the \textbf{natural multimodal human-computer interaction} experience.
We design additional state tokens, and corresponding training data and strategies to perceive various interaction scenarios.
The deployment of VITA employs a duplex scheme, where one model is responsible for generating responses to user queries, and the other continuously tracks environmental inputs, selectively outputting new responses with updated interactions.
This allows VITA to feature impressive human-machine interaction functionalities such as non-awakening interaction and audio interrupt interaction.
VITA is the \textbf{first step} for the open-source community to explore the seamless integration of multimodal understanding and interaction.
While there is still lots of work to be done on VITA to get close to close-source counterparts, we hope that its role as a pioneer can serve as a cornerstone for subsequent research. 
\end{abstract}

\section{Introduction}\label{sec:intro}

\begin{figure*}[t]
    \includegraphics[width=0.97\linewidth]{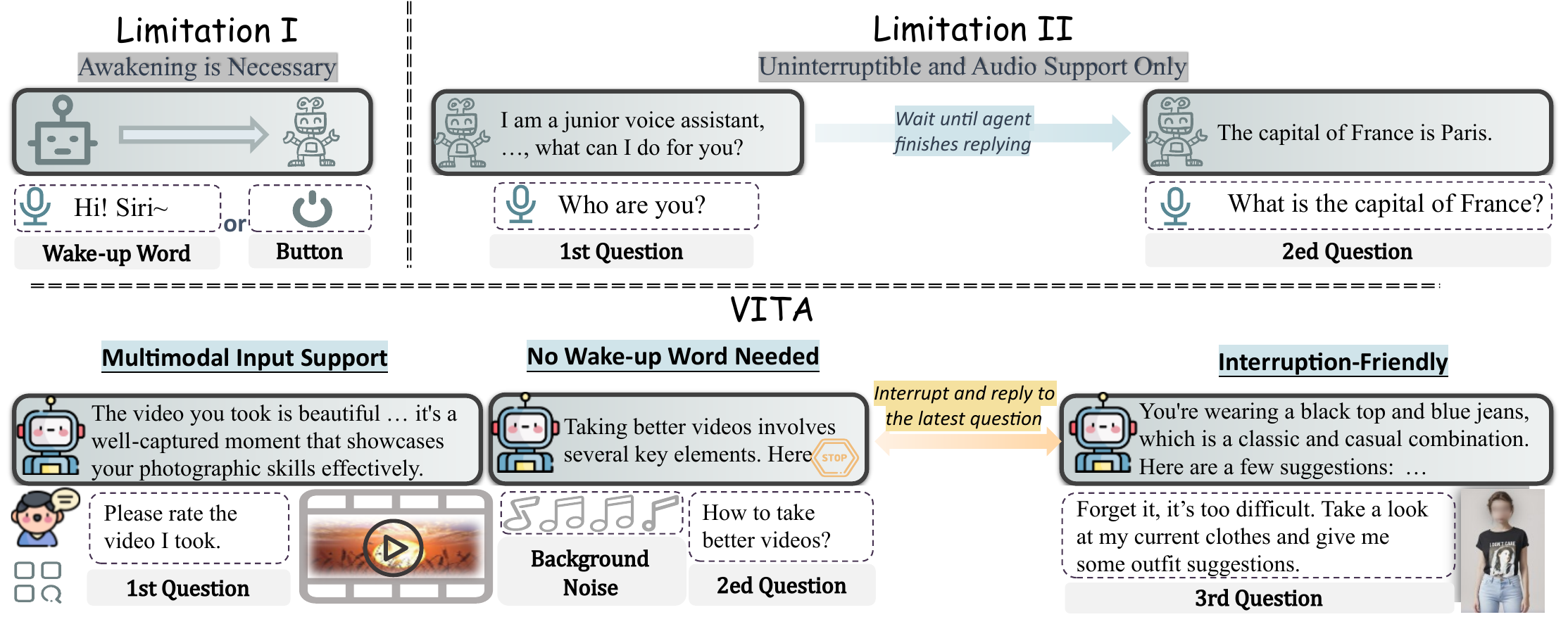}
 \caption{
 \textbf{Interaction of VITA}.
  Traditional audio interaction requires a pre-defined wake-up word, e.g., saying ``\textit{Hi! Siri$\sim$}'' every time you ask a question, or a button to control the input question audio \textbf{(Limitation 1)}.
  The human-computer interaction is always blocked when the model generates output, as the previous system can only respond to input queries sequentially \textbf{(Limitation 2)}.
  By contrast, on the one hand, unlike previous methods where audio, text, and vision are always separated, VITA supports these modalities end-to-end.
  On the other hand, VITA makes two contributions to multimodal interaction.
  \textbf{Non-awakening Interaction}: VITA automatically filters background noise like non-query human voices, thereby eliminating the need for the wake-up word and the button.
  \textbf{Audio Interrupt Interaction}: If the user interrupts with another question, the generation process is paused, and the model immediately responds to the latest query.
 }
\label{fig:development}
\end{figure*}

\begin{figure*}[h]
    \includegraphics[width=0.94\linewidth]{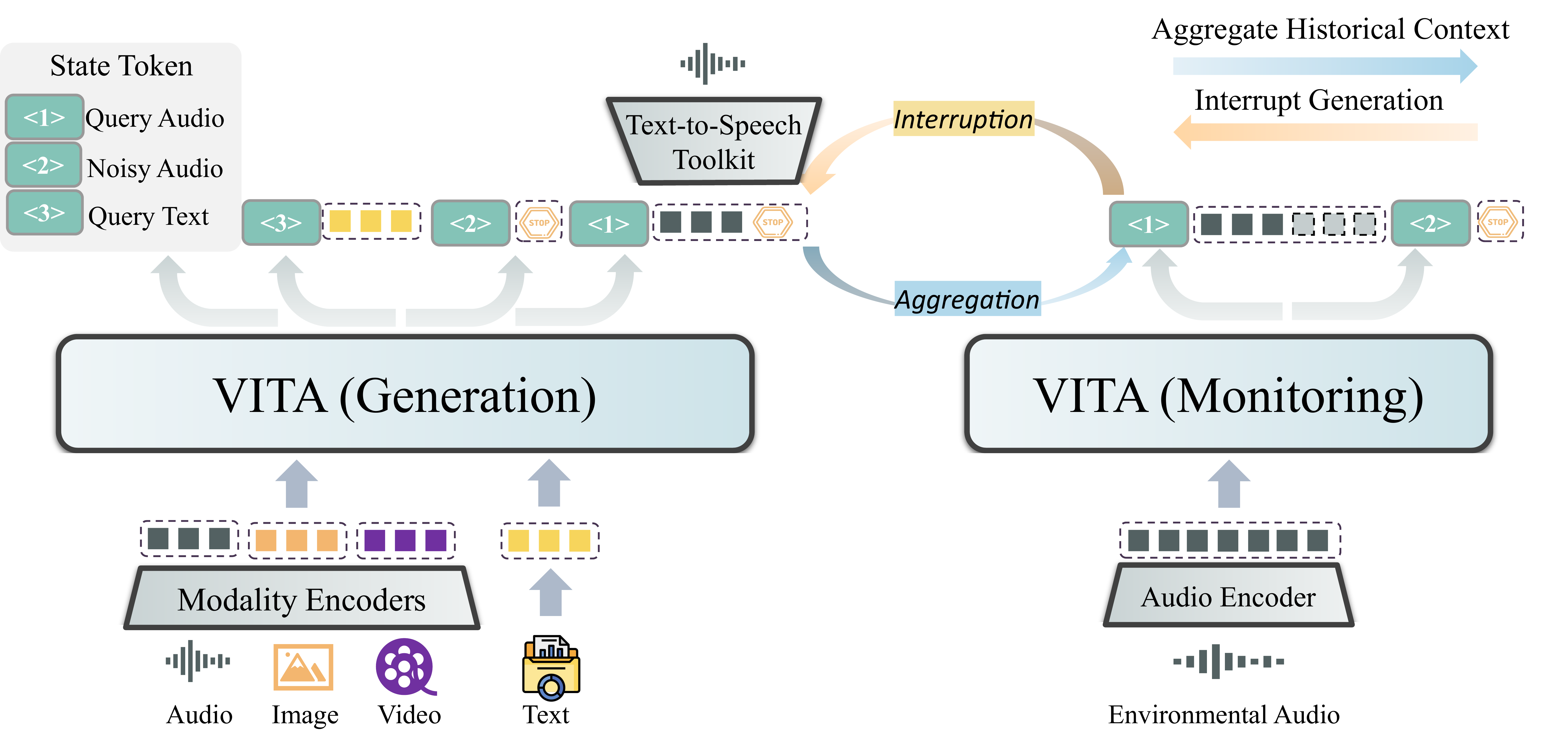}
 \caption{\textbf{Architecture of VITA}.
 VITA is capable of processing inputs in the form of pure text/audio, as well as video/image combined with text/audio.
 Besides, we set different state tokens for different query inputs.
 \texttt{<1>} corresponds to the effective query audio, such as ``what is the biggest animal in the world?'', for which we expect a response from the model.
 \texttt{<2>} corresponds to the noisy audio, such as someone in the environment calls me to eat, for which we expect the model not to reply.
 \texttt{<3>} corresponds to the query text, i.e., the question given by the user in text form.
 During the training phase, we try to teach the model to automatically distinguish different input queries.
 During the deployment phase, with \texttt{<2>} we can implement non-awakening interaction.
 Based on this, we further introduce a duplex scheme for the audio interrupt interaction.
 Two models are running at the same time, where the generation model is responsible for handling user queries.
 When the generation model starts working, the other model monitors the environment.
 If the user interrupts with another effective audio query, the monitoring model aggregates the historical context to respond to the latest query, while the generation model is paused and tune to  monitor, i.e., the two models swap identities.
 }
\label{fig:teaser}
\end{figure*}

Large Language Models (LLMs) have undergone significant evolution~\citep{gpt4, jiang2023mistral, chiang2023vicuna, touvron2023llama2, dubey2024llama} and rencently, we have witnessed the flourishing of Multimodal Large Language Models (MLLMs)~\citep{yin2023survey, li2024mini, liu2023visual, liu2024llavanext, gpt4o}, which exhibit surprising multimodal capabilities.
Particularly, GPT-4o~\citep{gpt4o} has significantly advanced the field of MLLMs, distinguishing itself with two key attributes:
(1) a unified framework that processes text, vision, and audio signals in an end-to-end manner, and (2) the capability to enable natural multimodal human-computer interaction.
These breakthroughs have established a new standard in the discipline.
However, there is a conspicuous absence of open-source models that echo these highlights.
The pressing need for open-source community to further promote development in this field cannot be overstated.

This paper introduces VITA, as a pioneer that has initially achieved the both two attributes, by integrating architectural innovations with advanced training and development strategies. 
The functionality and architecture of VITA are depicted in Fig.~\ref{fig:development} and Fig.~\ref{fig:teaser}, respectively.
The implementation of VITA mainly comprises three steps:

\textbf{Bilingual Instruction Tuning of LLM}.
The official Mixtral 8$\times$7B~\citep{jiang2024mixtral} always lacks proficient Chinese language expression and comprehension. To tackle this, we expand the vocabulary of the base model and continued with further instruction tuning using the collected high-quality bilingual text corpus.
This makes the LLM proficient in both Chinese and English.

\textbf{Multimodal Alignment and Instruction Tuning}.
To align the text feature space with video, image, and audio, we collect massive high-quality multimodal data to align individual encoders and connectors, which process different modalities respectively, to the LLM.
Multimodal instruction tuning data are meticulously constructed.
While giving VITA a powerful multimodal foundational capability, we teach it to recognize the type of input queries end-to-end by introducing a state token.
This makes it possible to interact without audio awakening while inference.

\textbf{Development with Duplex Pipeline}.
In terms of model deployment, we introduce a duplex scheme.
As shown in Fig.~\ref{fig:teaser}, two VITA models are deployed simultaneously: one is responsible for generating responses to the current audio query, and the other continuously monitors for the new one. 
If any, the current generation is interrupted, and the model outputs the response to the new query.
In order to improve the efficiency of the interaction, we have carried out massive engineering optimizations, such as adapting multimodal vLLM \citep{kwon2023efficient}.

The contributions of this paper are as follows:

\begin{itemize}
\item
We develop an open-source high-performance multimodal base model that simultaneously supports video, image, text, and audio inputs in both English and Chinese. The model accepts either pure text/audio inputs or video/image combined with text/audio inputs. 
We design a comprehensive training process, which includes enhancing the Chinese capabilities of the LLM, constructing multimodal training data, and a multi-stage training pipeline.

\item
As a pioneer, we make preliminary explorations in the field of natural multimodal human-computer interaction.
By introducing a state token, the model can automatically identify the type of the input audio to achieve non-awakening interaction. 
Meanwhile, the duplex scheme makes it possible to realize audio interrupt interaction.

\item
We fully open-source our model, training code, and inference deployment framework, aiming at promoting the advancements of the research community. As a cutting-edge research, we will continue to contribute to the multimodal foundation models and interactions.

\end{itemize}

\section{Related Work}
Leveraging advanced LLMs such as GPTs~\citep{gpt4,brown2020language}, LLaMA~\citep{touvron2023llama,touvron2023llama2}, Alpaca~\citep{taori2023stanford}, Vicuna~\citep{chiang2023vicuna}, and Mistral~\citep{jiang2023mistral}, MLLMs exhibit enhanced multimodal capabilities, particularly through end-to-end training techniques.
Recent open-source MLLMs, such as Otter~\citep{li2023otter}, mPLUG-Owl~\citep{ye2023mplug}, LLaVA~\citep{liu2023visual}, Qwen-VL~\citep{bai2023qwen}, Cambrian-1~\citep{tong2024cambrian}, Mini-Gemini~\citep{li2024mini}, MiniCPM-V 2.5~\citep{hu2024minicpm}, DeepSeek-VL~\citep{lu2024deepseek}, SliME~\citep{zhang2024beyond}, and Bunny~\citep{he2024bunny}, have made progress in solving multimodal fundamental problems, such as vision-language alignment and instruction following.

Among them, some representative open-source models like InternLM-XComposer-2.5~\citep{zhang2023internlm} and InternVL-2~\citep{chen2023internvl} have been rapidly advancing, demonstrating strong multimodal understanding capabilities and closely rivaling proprietary models in various multimodal benchmarks.
However, compared to proprietary models such as GPT-4o~\citep{gpt4o} and Gemini-Pro 1.5~\citep{team2023gemini}, which support more than two modalities like audio, image, and text, most open-source models focus on image-text modalities \citep{zhan2024anygpt}.
Furthermore, open-source models rarely concentrate on user interaction capabilities, leaving this area relatively unexplored.
In comparison, the proposed VITA not only exhibits impressive performance in perceiving data across four modalities, i.e., video, image, text, and audio, but also makes preliminary strides in enhancing user interaction capabilities.
Through the comprehensive open-sourcing of VITA, we hope to accelerate developments in this field.

\section{VITA}
\begin{figure*}
 \includegraphics[width=\linewidth]{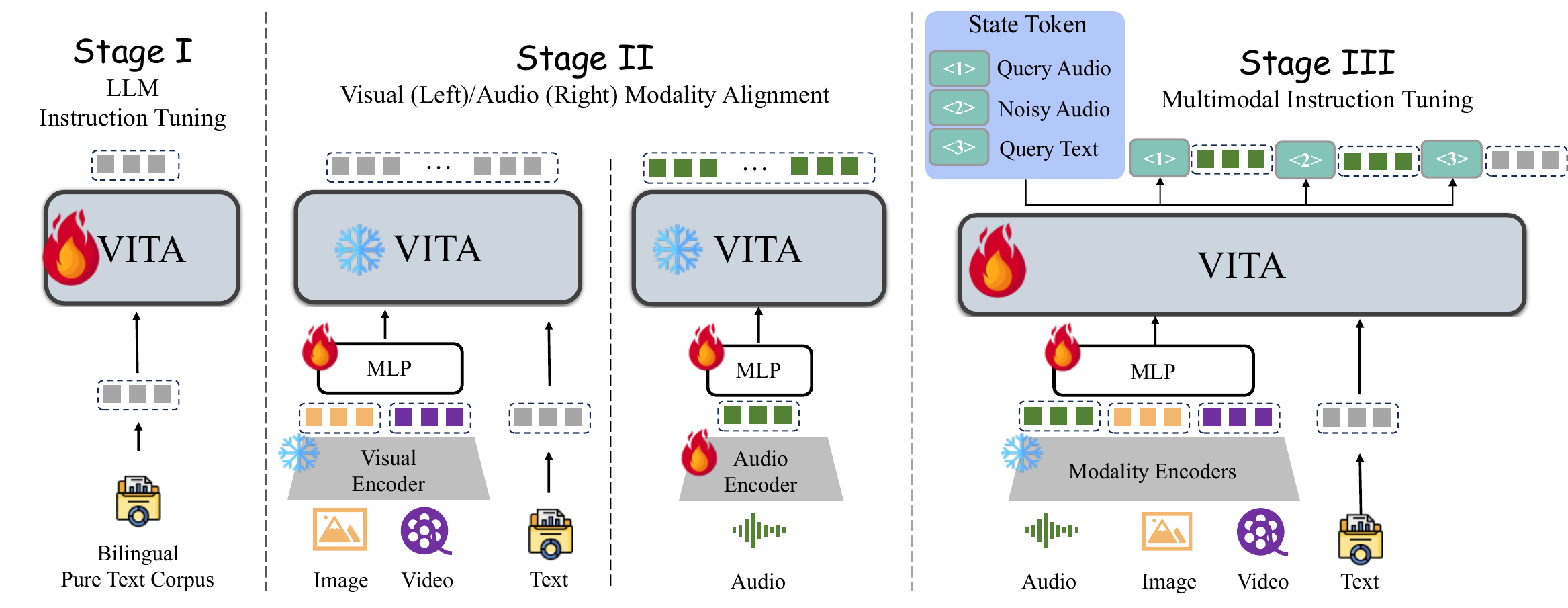}
    \caption{\textbf{Training pipeline of VITA}. The first stage \textbf{LLM Instruction Tuning} enhances the language model Mixtral 8$\times$7B by expanding its vocabulary size and fine-tuning it with a high-quality bilingual text corpus, thereby achieving proficiency in both Chinese and English. 
    The second stage \textbf{Multimodal Alignment} connects individual encoders with the LLM to process various modalities. By amassing a substantial collection of high-caliber multimodal data, we synchronize the text feature space with that of video, image, and audio.
    The last stage \textbf{Multimodal Instruction Tuning} allows the model to follow text or audio instructions to understand the image or video. A specially designed state token is used to distinguish the type of input query, facilitating subsequent multimodal human-computer interaction.}\label{fig:3stage_training}
\end{figure*}

As depicted in Fig.~\ref{fig:3stage_training}, the overall training pipeline of VITA consists of three stages: LLM instruction tuning, multimodal alignment, and multimodal instruction tuning.
The development of VITA is also an important part.

\subsection{LLM Instruction Tuning}

Mixtral 8x7B\footnote{\url{https://huggingface.co/mistralai/Mixtral-8x7B-v0.1}}~\citep{jiang2024mixtral} is a representative LLM with an architecture of sparse mixture of experts (SMoE).
Its performance is among the top-tier open-source LLMs, making it an ideal starting point of our work.
Nonetheless, we observe that the official Mixtral model exhibits limited proficiency in understanding Chinese.
To infuse bilingual (Chinese and English) comprehension capabilities, we broaden the vocabulary of the base model with Chinese, increasing the vocabulary size from $32,000$ to $51,747$.
This extension can also reduce the number of tokens under the same text, thus improving inference efficiency.
With the extended vocabulary in place, we use $5$ million synthetic bilingual corpus for pure-text instruction tuning.

\subsection{Multimodal Alignment} 
In this stage, we aim to bridge the representation gap between text and other modalities, thereby laying the groundwork for multimodal understanding.

\subsubsection{Visual Modality}

\begin{table}[ht]
    \centering
    \caption{Training data of multimodal instruction tuning. The images of the synthetic data come from open-source datasets Wukong~\citep{gu2022wukong}, LAION~\citep{schuhmann2022laion}, and CC12M~\citep{changpinyo2021conceptual}.}
    \label{tab:dataset_statistics}
    \resizebox{\textwidth}{!}{%
\begin{tabular}{@{}cccccccc@{}}
\toprule
\Gray
\textbf{Data Scenario} & \textbf{QA Type} & \textbf{Dataset Name} & \textbf{\begin{tabular}[c]{@{}c@{}}\# Concatenated \\ Entries (K)\end{tabular}} & \textbf{\begin{tabular}[c]{@{}c@{}}\# Total \\ Entries (K)\end{tabular}} & \textbf{\begin{tabular}[c]{@{}c@{}}\# Audio\\ Questions (K)\end{tabular}} & \textbf{\begin{tabular}[c]{@{}c@{}}\# Text \\ Questions (K)\end{tabular}} & \textbf{Language} \\ \hline
 &  & {ShareGPT4V} & 50.7 & 99.5 & 49.7 & 49.7 & Eng \\
 &  & {Allava-Caption} & 362.2 & 708.1 & 354.6 & 353.6 & Eng \\
 &  & {ShareGTP4o-Image} & 39.2 & 57.3 & 28.6 & 28.7 & Eng \\
 & \multirow{-4}{*}{\texttt{Description}} & {Synthetic Data} & 304.6 & 594.5 & 297.0 & 297.5 & CN \\ \cmidrule{3-8}
 &  & {LLaVA-150K} & 57.9 & 99.8 & 109.0 & 109.1 & CN \\
 &  & {LLaVA-Mixture-sample} & 308.5 & 308.5 & 1103.0 & 920.9 & Eng \\
 &  & {Lvis-Instruct} & 110.4 & 110.4 & 562.1 & 466.0 & Eng \\
 &  & {ScienceQA} & 12.7 & 12.7 & 0.0 & 12.7 & Eng \\
\multirow{-9}{*}{\texttt{General Image}} & \multirow{-5}{*}{\begin{tabular}[c]{@{}c@{}}\texttt{QA}\end{tabular}} & {Synthetic Data} & 14.1 & 21.8 & 106.0 & 105.3 & CN \\ \cmidrule{2-8}
 &  & {Anyword-3M} & 47.5 & 770.3 & 384.5 & 385.8 & CN \\
 &  & {ICDAR2019-LSVT} & 233.2 & 233.2 & 680.0 & 583.1 & CN \\
 &  & {ICDAR2017-RCTW} & 6.6 & 7.7 & 3.7 & 4.0 & CN \\
 &  & {Open-Chart} & 32.3 & 41.5 & 229.2 & 229.0 & CN \\
\multirow{-5}{*}{\texttt{OCR \& Diagram}} & \multirow{-5}{*}{\begin{tabular}[c]{@{}c@{}}\texttt{Description}\\ \&\\ \texttt{QA}\end{tabular}} & {Synthetic Data} & 97.3 & 156.0 & 418.8 & 345.2 & CN \\ \cmidrule{2-8}
 & \texttt{Description} & {ShareGemini} & 777.7 & 777.7 & 104.1 & 673.7 & CN \& Eng \\  \cmidrule{3-8}
\multirow{-2}{*}{\texttt{General Video}} & \texttt{QA} & {Synthetic Data} & 160.5 & 160.5 & 280.6 & 179.5 & CN \& Eng \\ \cmidrule{2-8}
\texttt{Pure Text} & \texttt{QA} & {Synthetic Data} & 134.5 & 800.8 & 398.5 & 1113.9 & CN \& Eng \\  \cmidrule{2-8}
& \multicolumn{2}{c}{\texttt{Total}} & 2749.9 & 4960.3 & 5109.4 & 5857.7 & CN \& Eng \\ \bottomrule
\end{tabular}%
}
\end{table}

\textbf{Visual Encoder}. 
We employ InternViT-300M-448px as the visual encoder\footnote{\url{https://huggingface.co/OpenGVLab/InternViT-300M-448px}}, which accepts a 448$\times$448 image as input, generating 256 tokens after using a visual connector that is a simple two-layer MLP.
For high-resolution image input, we implement the dynamic patching strategy~\citep{chen2024far} to capture local details.
Videos are treated as special cases of images.
If the video length is less than $4$ seconds, we uniformly sample $4$ frames.
If the video length is between $4$ and $16$ seconds, we sample one frame per second.
For videos longer than $16$ seconds, we uniformly sample $16$ frames.
To prevent the introduction of an excessive number of visual tokens, we do not perform dynamic patching on individual frames of the video.

\begin{wrapfigure}{r}{0.5\linewidth}
\vspace{-0.3cm}
  \begin{center}
    \includegraphics[width=\linewidth]{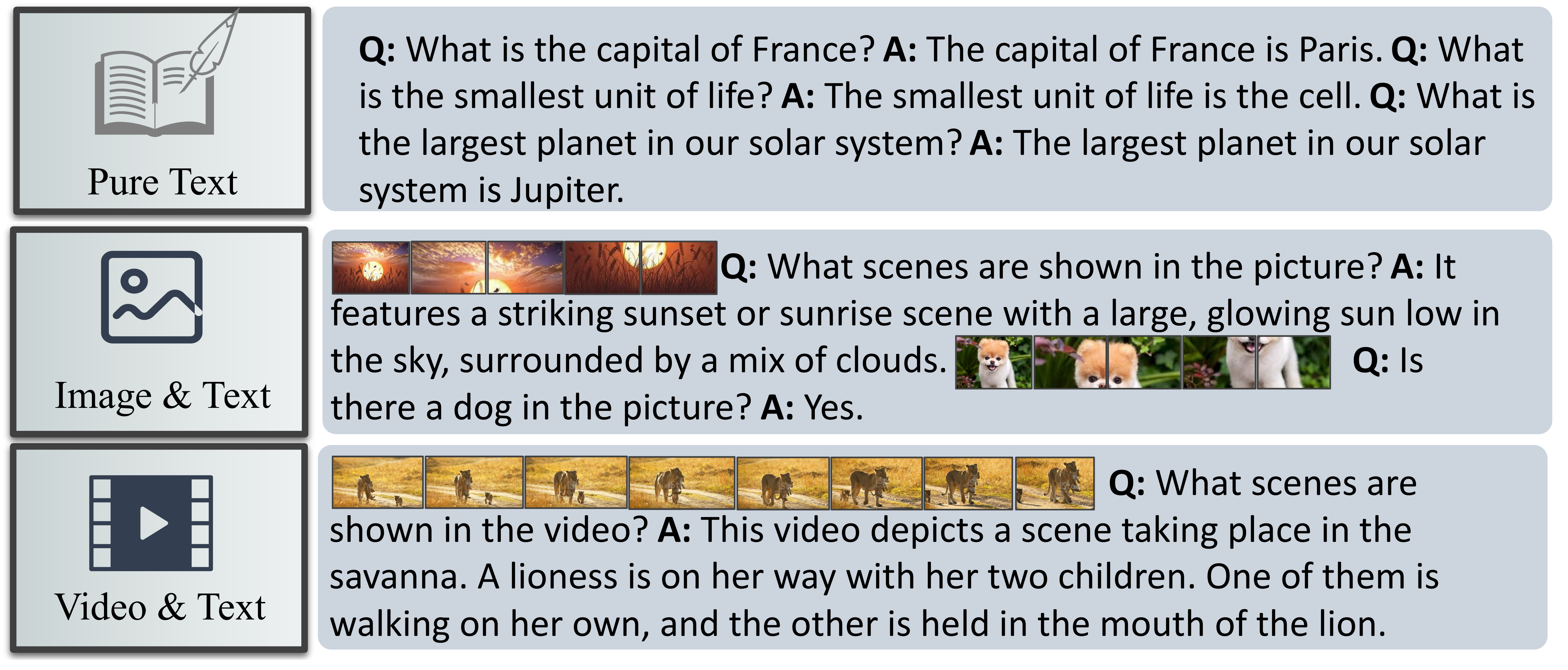}
\caption{\textbf{Data Concatenation.} Different text data is directly concatenated to 6K tokens. Images are first divided into local patches and then different image-text pairs are concatenated. Video data is directly sampled frame by frame as input, without the need for concatenation. In this way, we could unify the length of data in different training batches, thus improving training efficiency.} 
\label{fig:concate}
\end{center}
\vspace{-0.3cm}
\end{wrapfigure}
\textbf{Visual Alignment}.
We only train the visual connector during the visual alignment phase.
Table~\ref{tab:dataset_statistics} summarizes the used training data, except the pure text part.
In addition, in this phase we do not use audio questions.
(1) For the general image description task, we introduce the GPT-4V generated part from ShareGPT4V~\citep{chen2023sharegpt4v} to ensure data quality.
We also introduce Allava-Caption~\citep{chen2024allava} and ShareGTP4o-Image\footnote{\url{https://sharegpt4o.github.io/}}, and supplement these with Chinese image descriptions generated by existing MLLMs.
(2) For the general image Question-Answering (QA) task, we initially gather three datasets: LLaVA-Mixture-sample~\citep{liu2023visual}, Lvis-Instruct~\citep{wang2023instruct4v}, and ScienceQA~\citep{lu2022learn}.
We then use existing MLLMs to generate an additional 21.8K Chinese QA data. 
Besides, we remove the caption subset from LLaVA-150K~\citep{liu2023visual} and translate the rest into Chinese.
(3) For the OCR \& Diagram tasks, we include Anyword-3M~\citep{tuo2023anytext}, ICDAR2019-LSVT\footnote{\url{http://icdar2019.org/}}, ICDAR2017-RCTW\footnote{\url{https://iapr.org/icdar2017}}, Open-Chart (a collection of ChartQA~\citep{masry2022chartqa}, DVQA~\citep{kafle2018dvqa}, InfoVQA~\citep{mathew2022infographicvqa}, Pew~\citep{kantharaj2022chart}, and OpenCQA~\citep{kantharaj2022opencqa}), as well as some synthetic data generated by existing MLLMs from other open-source data with text.
For Anyword-3M, we select data where the answer's corresponding text length is between 20 and 50, with the question asking to identify the text in the image.

For ICDAR2019-LSVT, ICDAR2017-RCTW, and Open-Chart, we generate detailed descriptions and QA pairs using existing MLLMs.
(4) For general video descriptions, we use the ShareGemini~\citep{sharegemini} dataset.
(5) For general video QA, we re-label open-source data from Video-ChatGPT~\citep{Maaz2023VideoChatGPT} and VideoChat2~\citep{2023videochat} using existing MLLMs.
We sample $800$k entries from $5$ million pure text data to maintain the LLM's text comprehension abilities. It is used during multimodal instruction tuning instead of visual alignment, because the LLM parameters of the latter are frozen.

\textbf{Data Concatenation}. 
For pure text data and image data, we aim to concatenate the context length to 6K tokens as illustrated in Fig.~\ref{fig:concate} (the amount of concatenated data is indicated in the Concatenated Entries column of Table~\ref{tab:dataset_statistics}). 
Video data, on the other hand, is not subjected to concatenation. 
Concatenating different data offers two benefits: 
(1) It supports longer context lengths, allowing for the expansion from single to multiple image-question interactions, resulting in more flexible input forms and extended contexts. 
(2) It enhances computational efficiency, as video frames typically contain a high number of visual tokens. By concatenating image-question pairs, we maintain a balanced number of tokens in the training batch, thus improving computational efficiency. Furthermore, we find that the model trained with concatenated data perform comparably to those trained with the original data.

\subsubsection{Audio Modality}
\textbf{Audio Encoder}. 
The input audio is initially processed through a Mel Filter Bank block.
This block breaks down the audio signal into individual frequency bands on the mel frequency scale, mimicking the nonlinear human perception of sound.
Subsequently, we utilize 4$\times$CNN downsampling layers followed by a $24$ layers of transformer, totaling $341$M parameters, to process the input features.
We employ a simple two-layer MLP as the audio-text modality connector.
In the end, each $2$ seconds of audio input is encoded into $25$ tokens.

\textbf{Audio Alignment}.
For one of the alignment tasks, we have opted for Automatic Speech Recognition (ASR). Our dataset includes Wenetspeech~\citep{zhang2022wenetspeech}, which encompasses over 10,000 hours of multi-domain speech recognition data, with a primary focus on Chinese tasks. Similarly, Gigaspeech~\citep{chen2021gigaspeech} also contains 10,000 hours of high-quality audio data, with the majority of the data geared towards English speech recognition tasks. The other task is audio captioning, which relies on the AudioSet SL subset of Wavcaps~\citep{mei2024wavcaps}. This dataset features 400K audio clips along with their corresponding audio captions. During alignment, both the audio encoder and connector are trained.

\subsection{Multimodal Instruction Tuning}

During this stage, we perform instruction tuning on the model to enhance its instruction following capability, whether text or audio.

\subsubsection{Training Data}
\begin{table}[]
\caption{System prompts for {\setlength{\fboxsep}{0pt} \colorbox[HTML]{D9EAD3}{image input}}, {\setlength{\fboxsep}{0pt} \colorbox[HTML]{EAD1DC}{video input}}, and {\setlength{\fboxsep}{0pt} \colorbox[HTML]{CFE2F3}{pure text input}}.}\label{tab:sys_prompt}
\resizebox{\textwidth}{!}{%

\begin{tabular}{c}
\toprule
\rowcolor[HTML]{D9EAD3} 
\texttt{System prompt for image data}                                                                                                            \\
\begin{tabular}[c]{@{}l@{}}You are an AI robot and your name is VITA.\\ You are a multimodal large language model developed by the open-source community. Your aim is to be helpful, honest, and harmless.\\ You support the ability to communicate fluently and answer user questions in multiple languages of the user's choice.\\ If the user corrects the wrong answer you generated, you will apologize and discuss the correct answer with the user.\\ You must answer the question strictly according to the content of the image given by the user, and it is strictly forbidden to answer the question \\ without the content of the image. Please note that you are seeing the image, not the video.\end{tabular}  \\ \hline
\rowcolor[HTML]{EAD1DC} 
\texttt{System prompt for video data}                                                                                                                \\
\begin{tabular}[c]{@{}l@{}}You are an AI robot and your name is VITA. \\ You are a multimodal large language model developed by the open-source community. You aim to be helpful, honest, and harmless.\\ You support the ability to communicate fluently and answer user questions in multiple languages of the user's choice.\\ If the user corrects the wrong answer you generated, you will apologize and discuss the correct answer with the user.\\ You must answer the question strictly according to the content of the video given by the user, and it is strictly forbidden to answer the question \\ without the content of the video. Please note that you are seeing the video, not the image.\end{tabular} \\ \hline
\rowcolor[HTML]{CFE2F3} 
\texttt{System prompt for text data }                                                                                                              \\
\begin{tabular}[c]{@{}l@{}}You are an AI robot and your name is VITA.\\ You are a multimodal large language model developed by the open-source community. Your aim is to be helpful, honest, and harmless.\;\;\;\;\;\;\;\;\;\;\;\;\;\\ You support the ability to communicate fluently and answer user questions in multiple languages of the user's choice.\\ If the user corrects the wrong answer you generated, you will apologize and discuss the correct answer with the user.\end{tabular}                                                          \\ \bottomrule                                                                                                                                                                                   \end{tabular}
}
\end{table}

\textbf{Data Construction}.
The data source in the instruction tuning phase are same as the alignment phase in Table~\ref{tab:dataset_statistics}, and we make the following improvements:
(1) the questions are randomly (about half) replaced with their audio versions, using TTS technique such as GPT-SoVITS\footnote{\url{https://github.com/RVC-Boss/GPT-SoVITS}}, to enhance the model's understanding of audio queries and its instruction following capabilities. 
The number of audio questions and text questions can be found in Table~\ref{tab:dataset_statistics}.
(2) Different system prompts are set to avoid conflicts between different types of data, as listed in Table~\ref{tab:sys_prompt}.
For instance, some questions can be answered based on visual information or based on the model's own knowledge, leading to conflicts.
Additionally, since the image data have been patched that are similar to multiple frames of video data, which may confuse the model. 
The system prompt explicitly distinguishes different data types, making it more intuitive to understand.

\textbf{Noisy Audio Construction}. 
During human-computer interaction, not all audio inputs require a response, which are collectively referred to as noisy audio.
A system with good interactive capabilities should be able to actively identify the type of audio~\citep{dighe2024leveraging} and selectively execute subsequent outputs.
To this end, we need to construct various noisy audio samples for the model to recognize. 
Specifically, we randomly sample $474$K sentences from answers of existing multimodal and unimodal QA data.
These negative sample texts, focusing on non-query-related content that does not require a user response, have a length distribution consistent with the positive question length distribution. 
Then, we use the TTS tool to convert these sentences into audio.
The construction of noisy audio samples enables the model to recognize audio inputs that do not require a response, which is beneficial for implementing Non-awakening Interaction.
The specific training strategy will be elaborated in the following section.

\subsubsection{Training Process}\label{sec:sft_tune}
In accordance with the QA pairs constructed in the above section, the model needs to distinguish three types of queries:

- Query Audio: The question is initiated by audio.

- Noisy Audio: The input is audio, but it does not contain a question. 

- Query Text: The question is initiated by text.

Based on these query types, we have designed three state tokens \texttt{<1>}, \texttt{<2>}, and \texttt{<3>}.
During the training phase, we insert corresponding state tokens at the beginning of the answers, allowing the model to flexibly handle different interactive behaviors.
Specifically:

- State token \texttt{<1>} denotes that the question input is the query audio. 
In this case, the output of the model needs to be presented to the user, either as text or speech converted by TTS tools.

- State token \texttt{<2>} indicates that the question input is the noisy audio.
The model should output an EOS token as a terminator.
However, we observe that abruptly terminating the output during training can significantly degrade performance.
Consequently, we send the text corresponding to the noisy audio to a LLM and use its output text as the training target.
During inference, \texttt{<2>} serves as another special EOS token.

- State token \texttt{<3>} signifies the question of pure text, which is used to distinguish between the above two queries in the training set.

During training, both visual and audio encoders are frozen, and the connectors are trained in conjunction with Mixtral 8$\times$7B.

\subsection{Development with Duplex Pipeline}
In this section, we primarily discuss how we implement two interaction functionalities, namely non-awakening interaction and audio interrupt interaction.

\subsubsection{Non-awakening Interaction}
Non-awakening interaction implies that the model can be activated and respond to user audio questions in the environment without the need for a wake-up word or button.
The deployment process must meet the following requirements:

- Real-time Tracking of Environmental Sounds. This involves determining whether the audio content constitutes human speech.

- Filtering out noisy audio. The model should only respond to effective human query audio.

For the first requirement, existing Voice Activity Detection (VAD) can provide assistance. It is also known as speech activity detection or speech detection, identifying the presence of human speech.
VITA employs SileroVAD~\citep{Silero_VAD}, which is trained on huge corpora that include over $6,000$ languages and performs well with various background noise.  
For the second requirement, we leverage the state token \texttt{<2>} described in Sec.~\ref{sec:sft_tune}.
This allows the model to automatically distinguish whether the input audio is an effective query.
If the input is of a non-query type, the model directly terminates the inference, thereby only responding to query-type inputs.

\subsubsection{Audio Interrupt Interaction}
Audio interrupt interaction enables users to interrupt the model's generation at any time with new questions.
To accomplish this, the deployment environment must fulfill the following requirements:

- Real-time Tracking and Filtering of External Queries. While generating responses, the system must simultaneously track and filter external queries in real time.

- Answering New Questions. When a new question emerges, the system must cease its current generation, consolidate the historical context, and respond to the present query.

To achieve this, we propose the duplex deployment framework, which is also an important research direction in audio field~\citep{ma2024language}.
As illustrated in Fig.~\ref{fig:development}, two VITA models are deployed concurrently.
Under a typical condition, the Generation model answers user queries.
Simultaneously, the Monitoring model detects environmental sounds during the generation process.
It disregards non-query user sounds, i.e., noisy audio, but ceases the Generation model's progress when it identifies query audio.
The Monitoring model subsequently consolidates the historical context and responds to the latest user query.
At this point, the identities of the Generation model and the Monitoring model are transformed.

\section{Evaluation}

\begin{table}[]
\centering
\caption{
Comparison of official Mixtral 8x7B Instruct and our trained Mixtral 8x7B.
``CN''/``ENG'' denote that the benchmark contains Chinese/English data.}\label{tab:pt}
\resizebox{0.7\linewidth}{!}{%
\begin{tabular}{lcccc}
\toprule
\multicolumn{1}{c}{\multirow{2}{*}{{Method}}} & {C-EVAL} & {AGIEVAL} & {MMLU} & {GSM8K} \\  \cmidrule{2-5}
\multicolumn{1}{c}{} & {CN} & {CN \& ENG} & {ENG} & {ENG} \\ \midrule
Mixtral-8x7B Instruct & 53.30 & 41.72 & 70.35 & 63.99 \\ \Gray
Mixtral-8x7B Ours & 56.68 & 46.17 & 70.98 & 75.66 \\ \bottomrule
\end{tabular}%
}
\end{table} 

\textbf{Language Performance.}
To validate the efficacy of our training process for language model, we evaluate our trained model ``Mixtral 8x7B Ours'' against the official version ``Mixtral 8x7B Instruct'', on four datasets: C-EVAL~\citep{huang2024c}, AGIEVAL~\citep{zhong2023agieval}, MMLU~\citep{hendrycks2020measuring}, and GSM8K~\citep{cobbe2021training}.
These datasets encompass a variety of scenarios including general multiple-choice questions, multidisciplinary QA, as well as mathematical and logical reasoning tasks, covering both Chinese and English contexts.
The results presented in Table~\ref{tab:pt} demonstrate that our training significantly enhances the language model's capabilities on Chinese evaluation sets (C-EVAL and AGIEVAL), while maintaining original performance levels on the English related benchmark (MMLU) and showing notable improvement in the mathematical reasoning task (GSM8K).

\begin{table}[]
\caption{Evaluation on ASR tasks. ``CN''/``ENG'' refers to Chinese/English speech. The metric of wenetspeech/librispeech is CER (Character Error Rate)/WER (Word Error Rate).}\label{tab:audio}
\centering
\resizebox{0.9\linewidth}{!}{%
\begin{tabular}{lcccccccc}
\toprule
\multirow{2}{*}{Method}& \textbf{} &\multicolumn{2}{c}{{Wenetspeech (CN)}} & \textbf{} & \multicolumn{4}{c}{{Librispeech (ENG)}} \\
\cmidrule{3-4}\cmidrule{6-9}
 && {Test\_Net} & {Test\_Meeting} & \textbf{} & {Dev\_clean} & {Dev\_other} & {Test\_clean} & {Test\_other} \\ 
\cmidrule{1-1}\cmidrule{3-4}\cmidrule{6-9}
VITA           && 12.15 & 16.53 &  & 7.57 & 16.57 & 8.14 & 18.41 \\
\bottomrule
\end{tabular}%
}
\end{table}

\textbf{Audio Performance.}
To validate the robustness of the speech representations learned by our model, we test it on the Wenetspeech\footnote{\url{https://github.com/wenet-e2e/WenetSpeech}} and Librispeech\footnote{\url{https://www.openslr.org/12}} datasets.
The Wenetspeech features two evaluation splits: test\_net and test\_meeting.
The former has data sources that are more closely aligned with the training data, making it easier, while the latter presents a greater challenge.
As a held-out dataset for our model, Librispeech assesses the model's generalization ability on unseen datasets.
It has four evaluation splits: those starting with ``dev'' are validation sets, and those starting with ``test'' are test sets.
``Clean'' refers to less challenging sets, while ``other'' indicates more challenging ones.
We can see that VITA has achieved considerable results on the ASR benchmarks.

\begin{figure}
    \centering
    \includegraphics[width=\linewidth]{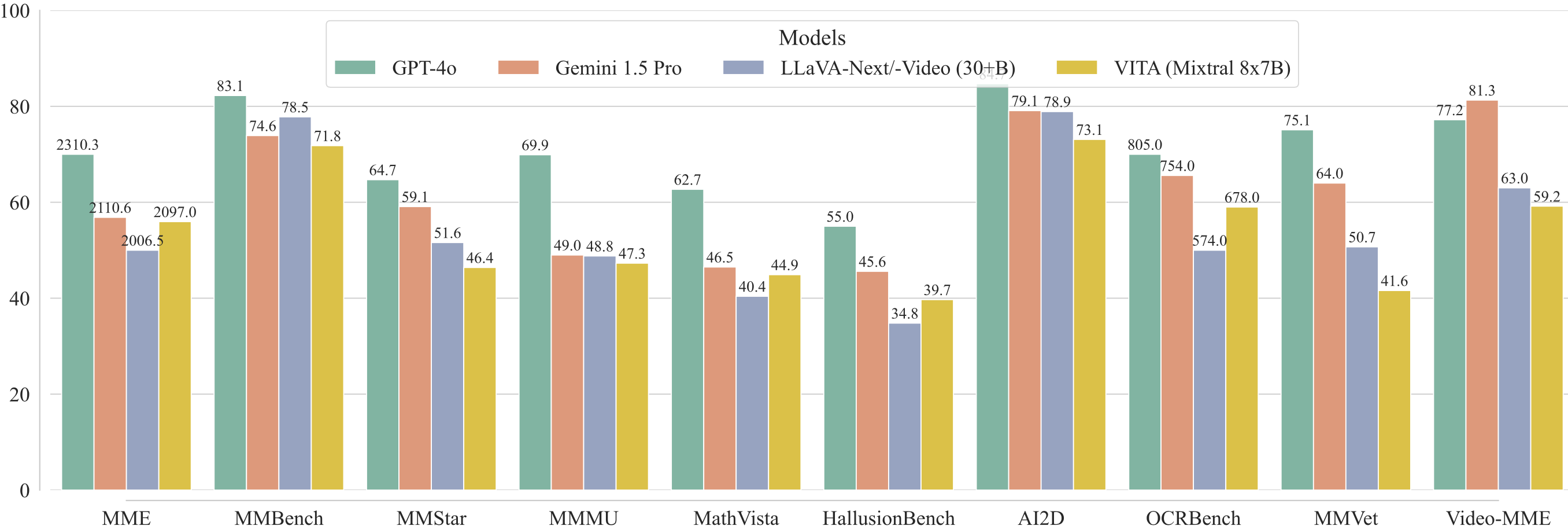}
    \caption{Evaluation on image and video understanding. VITA exhibits comparable performance to the current leading open-source models, but still fell short of advanced closed-source counterparts.}
    \label{fig:image_video}
\end{figure}

\textbf{Multimodal Performance.}
To assess multimodal capabilities, we evaluate VITA on ten representative benchmarks, including MME~\citep{fu2023mme}, MMBench~\citep{liu2023mmbench}, MMStar~\citep{chen2024we}, MMMU~\citep{yue2024mmmu}, MathVista~\citep{lu2023mathvista}, HallusionBench~\citep{guan2024hallusionbench}, AI2D~\citep{hiippala2021ai2d}, OCRBench~\citep{liu2023hidden}, MMVet~\citep{yu2023mm}, and Video-MME~\citep{fu2024video}.
The last one is a video benchmark, and the others belong to image evaluation sets. 
As depicted in Fig.~\ref{fig:image_video}, in terms of image understanding, VITA shows comparable performance with image specialized open-source model LLaVA-Next~\citep{liu2024llavanext} and is close to closed-source model Gemini 1.5 Pro~\citep{team2023gemini}.
In video understanding, although there is a small gap between VITA and the video-specialized LLaVA-Next-Video~\citep{zhang2024llavanext-video}, this is acceptable given that VITA supports a broader range of modalities and prioritizes interaction.
However, it is worth noting that a substantial gap still exists between current open-source models and proprietary models.

\section{Conclusion and Future Work}\label{sec:conclusion}
\vspace{-0.1cm}

In this paper, we have introduced VITA, a strong open-source MLLM that integrates video, image, text, and audio understanding into a unified framework, with advanced interactive experience.
Apart from robust multimodal fundational capabilities, VITA pioneers novel multimodal interactions for the open-source community, through non-awakening interaction and audio interrupt interaction.
However, the current version still has the following limitations:

- Enhancement of Foundational Capabilities.
While VITA demonstrates competitive performance in unimodal and multimodal tasks relative to leading open-source models, there remains a notable gap compared to proprietary counterparts.

- Refinement of Noisy Audio Construction.
Using non-query responses of existing data as noisy audio samples is simple yet effective.
However, there are instances where VITA misclassifies noisy audio as query audio, highlighting the need for a more nuanced construction approach.

- Building end-to-end TTS in conjunction with LLM.
We currently use an additional TTS tool to convert LLM generated text into speech, which is quite time-consuming. If TTS can be combined with LLM to achieve end-to-end speech output, it may greatly boost the real-time interaction.

\bibliographystyle{plain}
\bibliography{neurips_2024}

\clearpage
\newpage
\appendix

\end{document}